\documentclass{article}

\usepackage[preprint,nonatbib]{neurips_2024}

\usepackage[utf8]{inputenc} 
\usepackage[T1]{fontenc}    
\usepackage{hyperref}       
\usepackage{url}            
\usepackage{booktabs}       
\usepackage{amsfonts}       
\usepackage{nicefrac}       
\usepackage{microtype}      
\usepackage{xcolor}         

\usepackage{color}
\usepackage{colortbl}
\usepackage{wrapfig,lipsum,booktabs}

\usepackage{graphicx}
\usepackage{algorithm,algpseudocode}
\usepackage{amssymb}

\makeatletter
\newcommand\footnoteref[1]{\protected@xdef\@thefnmark{\ref{#1}}\@footnotemark}
\makeatother

\newcommand{\shorteq}{%
  \settowidth{\@tempdima}{-}
  \resizebox{\@tempdima}{\height}{=}%
}

\mathchardef\mhyphen="2D

\definecolor{RowColorCode}{rgb}{0.61,0.57,0.89}

\title{Financial Models in Generative Art: Black-Scholes-Inspired Concept Blending in Text-to-Image Diffusion}

\author{%
 Divya Kothandaraman, Ming Lin, Dinesh Manocha\\
  Department of Computer Science\\
  University of Maryland College Park, USA \\
}

\begin{document}

\maketitle

\begin{abstract}
    We introduce a novel approach for concept blending in pretrained text-to-image diffusion models, aiming to generate images at the intersection of multiple text prompts. At each time step during diffusion denoising, our algorithm forecasts predictions w.r.t. the generated image and makes informed text conditioning decisions. Central to our method is the unique analogy between diffusion models, which are rooted in non-equilibrium thermodynamics, and the Black-Scholes model for financial option pricing. By drawing parallels between key variables in both domains, we derive a robust algorithm for concept blending that capitalizes on the Markovian dynamics of the Black-Scholes framework. Our text-based concept blending algorithm is data-efficient, meaning it does not need additional training.  Furthermore, it operates without human intervention or hyperparameter tuning. We highlight the benefits of our approach by comparing it qualitatively and quantitatively to other text based concept blending techniques, including linear interpolation, alternating prompts, step-wise prompt switching, and CLIP-guided prompt selection across various scenarios such as single object per text prompt, multiple objects per text prompt and objects against backgrounds. Our work shows that financially inspired techniques can enhance text-to-image concept blending in generative AI, paving the way for broader innovation. Code is available at https://github.com/divyakraman/BlackScholesDiffusion2024.

\end{abstract}

\begin{figure}
    \centering
    \vspace*{-1em}
    \includegraphics[width=0.99\textwidth]{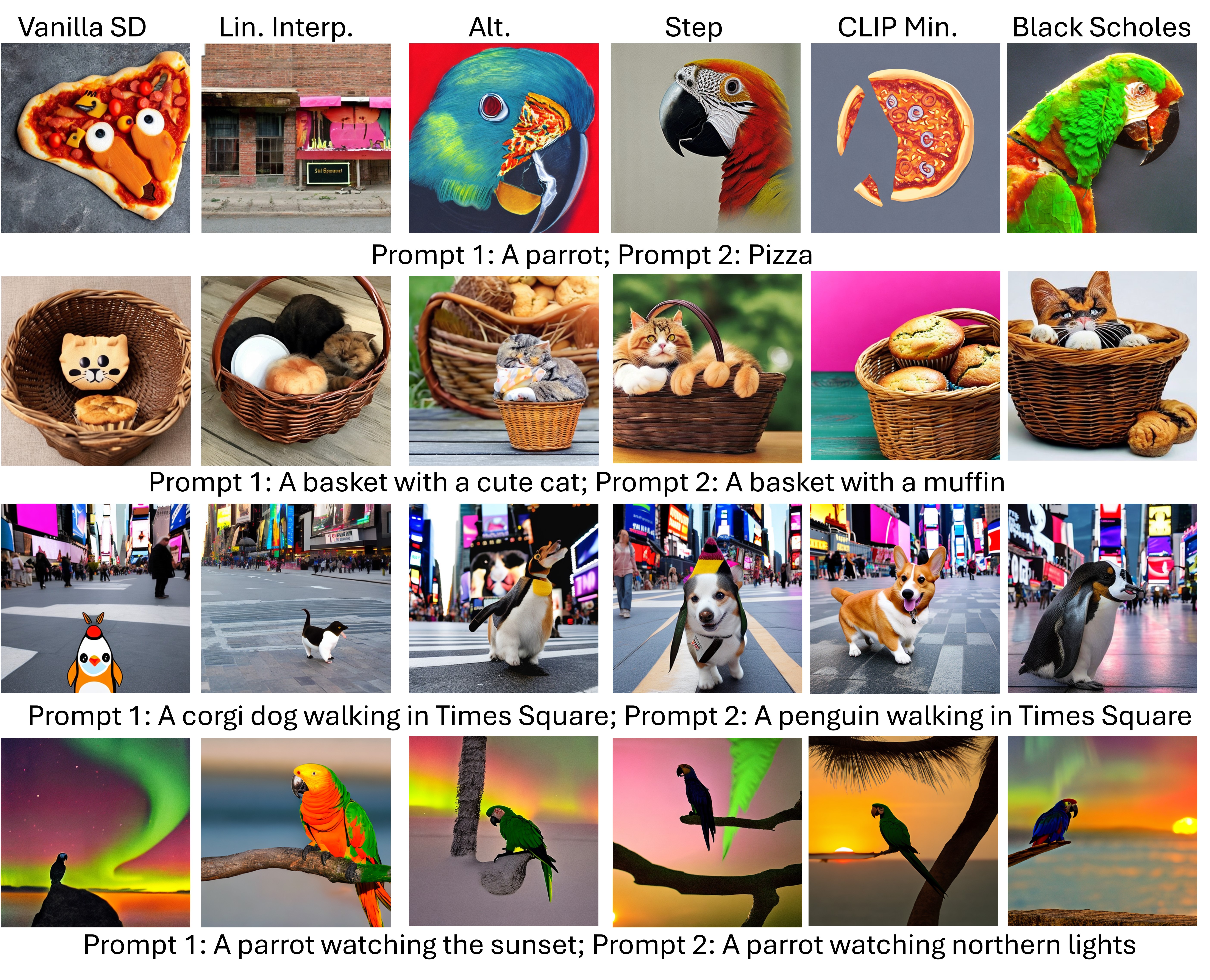}
    \vspace*{-1em}
    \caption{Our method’s results ({\bf Black Scholes}, last column) are presented alongside comparisons to prior work. Vanilla stable diffusion (SD) struggles to capture clear characteristics of individual text prompts (notably missing distinct features such as those of the parrot, cat, dog/penguin, and sunset/penguin). Linear interpolation performs poorly due to non-linear manifolds. Alternating sampling and step-wise switching yield low-quality results with artifacts, primarily because they lack intelligent prompt selection during denoising steps (missing characteristics of pizza, artifacts in cat/muffin and dog/penguin mixing, sunset/northern lights not well captured). CLIP-min exhibits bias issues by not modeling diffusion denoising dynamics and prompt selection effectively, which hinders fore-sighted decision making, the generated images are biased towards one of the text prompts. In contrast, our Black-Scholes model generates realistic images that meticulously balance and preserve the characteristics of each individual text prompt. The images are from set 1, set 2, set 3 and set 4 (Refer Section~\ref{sec:data}) respectively.}
    \label{fig:main1}
    \vspace*{-1em}
\end{figure}

\section{Introduction}

Text-to-image diffusion models~\cite{ho2020denoising,ho2022cascaded} have become essential tools for creative applications such as content creation, editing, personalization, and zero-shot generation. However, a core challenge remains: generating new and unseen concepts without retraining the model or collecting additional task-specific data. One promising avenue is to leverage the model’s intrinsic understanding of various conceptual manifolds to synthesize novel outputs by operating at their intersection.

This process, known as \emph{concept blending} or \emph{prompt mixing}~\cite{patashnik2023localizing,balaji2022ediff,CLIPGuidedImageMixing,imagemixerdiffusion}, used interchangeably in this paper, involves integrating two distinct concepts to produce a single, coherent image. For example, one could generate an image that merges the features of a pink cat with those of a dog. Although state-of-the-art diffusion models~\cite{adobefirefly,saharia2022photorealistic,betker2023improving,esser2024scaling} can blend text prompts into visually appealing results, these capabilities typically depend on advanced architectures and large-scale training data.

In contrast, data-efficient prompt mixing techniques hold the potential to enhance image generation while avoiding extra data collection costs. While simple strategies like linear interpolation of text embeddings~\cite{kawar2023imagic} exist, they may not be optimal given the highly nonlinear nature of the text-image manifold and the risks of bias. Other methods, such as alternating~\cite{kothandaraman2023aerial} or step-wise prompt switching~\cite{patashnik2023localizing}, encounter challenges stemming from human involvement and meticulous prompt engineering.

The central question we address is: \emph{How can we optimally switch between text prompts for effective concept blending?} An ideal solution would \textbf{automatically account for the varying capabilities} of the model with respect to different prompts. While attention maps~\cite{hong2023improving,chefer2023attend} and layout guidance~\cite{zheng2023layoutdiffusion,cheng2023layoutdiffuse} excel at directing the model toward distinct scene entities, they are less effective for blending concepts within the same entity. To overcome this, we propose an automated mechanism in which, at each step of the diffusion denoising process, the model focuses on the text prompt most aligned with the current image deficiencies.

\vspace*{-1em}
\paragraph{Main Contributions.} In this paper, we introduce a novel approach for concept blending that \textit{incorporates ideas from economics and finance}. Our method leverages pre-trained diffusion models to generate images that reside at the intersection of various text-image manifolds, dynamically conditioning the generation process on the prompt demanding the greatest attention. Specifically, our algorithm evaluates the ``cost'' associated with each text prompt and selects the optimal conditioning for the subsequent generation step based on which prompt requires the most optimization effort.

\textbf{Key Insight:} Our approach is inspired by \emph{non-equilibrium thermodynamics}~\cite{sohl2015deep}—the foundational principle behind diffusion models—and shares a conceptual basis with the Black--Scholes model~\cite{merton1976option,black2019pricing}, a celebrated framework in financial markets for pricing European stock options. In our analogy, the image functions as a valuable asset whose generation should occur at the most favorable ``cost,'' determined by its alignment with the input text prompts. Concretely, at each denoising step we:
\begin{enumerate}
    \item \textbf{Extrapolate the latent representations} to compute analogous ``stock prices'' at each timestep.
    \item \textbf{Leverage structural properties} of diffusion models to derive the variables in the Black--Scholes algorithm.
    \item \textbf{Predict a score} for each text prompt, indicating how the image should be conditioned in subsequent steps.
\end{enumerate}

This mechanism enables our model to adaptively concentrate on the aspects most in need of refinement, ultimately generating images that optimally reflect all relevant text inputs.

We validate our approach through extensive experiments on prompts of varying complexity, demonstrating the model's ability to seamlessly blend different objects and backgrounds. Quantitative and qualitative comparisons against baselines such as vanilla Stable Diffusion, linear interpolation, alternating sampling, prompt switching, and CLIP-guided prompt switching confirm the superiority of our method over current state-of-the-art techniques. Our work demonstrates that integrating financial methodologies into text-to-image synthesis can significantly improve concept blending, laying the groundwork for broader innovations in generative AI.

\section{Related work}

\subsection{Concept Blending / Prompt Mixing}
Prompt engineering~\cite{witteveen2022investigating} employs techniques such as rephrasing prompts to enhance model generalization. Additionally, large language models (LLMs)~\cite{lian2023llm,wu2023self} have been used to parse complex prompts and extract useful priors for image generation. Complementary to these methods are techniques involving attention maps~\cite{chefer2023attend} and prompt mixing, which applies different text prompts at various steps during the diffusion denoising process. For instance, Patashnik et al.~\cite{patashnik2023localizing} adopt a step-wise approach, whereas Aerial Diffusion~\cite{kothandaraman2023aerial} alternates prompts to produce semantically consistent aerial-view images. Similar strategies are employed by tools such as Image Mixer Diffusion~\cite{imagemixerdiffusion} and CLIP Guided Image Mixing~\cite{CLIPGuidedImageMixing}. A common drawback is that these methods often necessitate complex hyperparameter tuning.

\subsection{Mixing Step}
The \emph{mixing time} (or \emph{mixing step}) of a Markov chain, as discussed in~\cite{levin2017markov}, represents the time required for the chain to reach its steady-state distribution. This concept has been exploited in areas such as image editing~\cite{zhu2024boundary} and the synthesis of out-of-distribution (OOD) images~\cite{zhu2023unseen}. When using mixing time for prompt switching—similar to the step-wise methods of Patashnik et al.~\cite{patashnik2023localizing}—one key advantage is that the optimal switching time is determined mathematically, thereby reducing the need for extensive hyperparameter tuning. Typically, the mixing time is approximated by estimating the radius of the latent space. However, these approaches often apply to diffusion models trained on relatively small, problem-specific datasets and may struggle with the challenges of larger-scale foundation models. Moreover, they generally lack the flexibility to select the most optimal prompt at every timestep.

\subsection{Understanding the Latent Space}
Exploration of the latent space has been a significant area of investigation~\cite{karras2017progressive,abdal2019image2stylegan,gal2022stylegan}. Insights from these studies have facilitated advancements in downstream tasks such as image editing and manipulation~\cite{zhu2016generative,shen2020interpreting,kwon2022diffusion,zhu2020domain,preechakul2022diffusion}. Furthermore, a deeper understanding of the latent space in diffusion models has spurred improvements in various methods~\cite{rombach2022high,kwon2022diffusion,yang2023diffusion}, providing a foundation for more effective image synthesis and manipulation techniques.

\section{The Black Scholes Algorithm and Diffusion Models}

\subsection{The Black Scholes Model}

\label{sec:prelim_bs}

In this section, we provide a brief overview of the Black Scholes pricing model~\cite{merton1976option} used to determine the price of European call options of assets. In simple terms, a European call option allows an investor to lock in the price of an asset at any time but permits stock purchase (if desired) only upon expiration. Regardless of whether the stock price moves favorably or unfavorably over time, this option structure remains consistent. Investors rely on the Black-Scholes model to predict stock prices over time and make informed decisions about the optimal timing for stock purchases. The Black Scholes formula involves $5$ key variables: 
\begin{enumerate}
  \item Underlying stock price or spot price $S$: This represents the current price of the asset. 
  \item Strike price $K$: The strike price is the cost of the asset at the time of expiry. 
  \item Time to expiration $t$: It measures the time difference between the current moment and the expiry time. 
  \item Volatility $\sigma$: Volatility reflects the variation in prices of the asset. 
  \item Risk free rate $r$: The risk-free rate is the minimum return on an investment when the investor faces zero risks.
\end{enumerate}

To obtain the \textbf{Black Scholes score} of purchasing an asset, the spot price $S$ is first multiplied by the standard normal probability distribution function. From this result, to obtain the final cost $C$, the strike price $K$ multiplied by the cumulative standard distribution function is subtracted. Mathematically,
\begin{equation}
    SN(d_{1}) - Ke^{-rt}N(d_{2}), 
    \label{eq:bs}
\end{equation}
where
\begin{equation}
    d_{1} = \frac{log\frac{S}{K} + (r + \frac{\sigma^{2}}{2})(t)}{\sigma\sqrt{t}}, d_{2} = d_{1} - \sigma\sqrt{t}.
    \label{eq:bs_d}
\end{equation}

\subsection{Relation to Diffusion Models} 

\subsubsection{Diffusion Models}

The main concept behind diffusion models involves iteratively adding small amounts of random Gaussian noise to transform an initial photorealistic image \(x_{0}\) into noise \(x_{T} \sim \mathcal{N}(0,I)\) over \(T\) steps. This process is known as the \textbf{forward process}. Conversely, starting from random noise \(x_{T} \sim \mathcal{N}(0,I)\) and refining it iteratively for \(T\) steps can generate a photorealistic image \(x_{0}\). Since diffusion is gradual, \(T\) is typically large. At each intermediate timestep \(t \in \{0, \ldots, T\}\), \(x_{t}\) satisfies:

\[ x_{t} = \sqrt{\alpha_{t}}x_{0} + \sqrt{1 - \alpha_{t}}\epsilon_{t} \]

The hyperparameters of the diffusion schedule are \(0 = \alpha_{T} < \alpha_{T-1} < \ldots < \alpha_{1} < \alpha_{0} = 1\), and \(\epsilon_{T} \sim \mathcal{N}(0,I)\). To obtain \(x_{t-1}\) at each refinement step, the neural network \(f_{\theta}(x_{t}, t)\) is applied along with the corresponding random Gaussian noise perturbation. Essentially, during each step of diffusion denoising, the known variance in added noise follows a Gaussian distribution.

\subsubsection{Analysis based on Thermodynamics}

The mathematical formulation of diffusion models, which is rooted in non-equilibrium thermodynamics, exhibits striking similarities to the derivation of the Black-Scholes model used for pricing European call options in financial markets. Both models arise from analogous assumptions and share a common mathematical structure.

\paragraph{Diffusion Models}

In the context of diffusion, tools from statistical mechanics reveal that the generative dynamics undergo phase transitions and symmetry breaking. The dynamic equation of the generative process can be interpreted as a stochastic adiabatic transformation that minimizes free energy while maintaining thermal equilibrium. Moreover, diffusion models incorporate Gaussian noise in a controlled manner to learn the underlying data distribution: forward process, driving the system out of equilibrium by adding noise and reverse process, guiding the system back to a structured state.

\paragraph{The Black-Scholes Model}

The Black-Scholes model assumes that the stock price \(S_t\) follows a geometric Brownian motion, described by a stochastic differential equation (SDE). With the application of Itô’s Lemma, one derives a partial differential equation (PDE) that characterizes the evolution of option prices. Here, the pricing of derivatives can be viewed as a process of minimizing a form of ``financial free energy'' under specific constraints, analogous to thermodynamic systems seeking equilibrium.

\paragraph{A Unified Thermodynamic View}

Both models can be interpreted through the lens of thermodynamics:

\begin{itemize}
    \item In diffusion models, the reverse process minimizes a free energy functional:

\[
    F[p] = \int p(\mathbf{x}) \left( \log p(\mathbf{x}) - \log q(\mathbf{x}) \right) d\mathbf{x},
    \]

    where \(q(\mathbf{x})\) represents the target distribution.
    
    \item In the Black-Scholes framework, option pricing is analogous to minimizing a financial free energy, where the equilibrium state corresponds to a fair, risk-neutral price.
\end{itemize}

Thus, while diffusion models and the Black-Scholes equation address different applications—one in generative modeling and the other in finance—they both embody the principles of free energy minimization and equilibrium. The forward process in diffusion models perturbs the system away from equilibrium using controlled noise, whereas the Black-Scholes approach interprets market fluctuations through Brownian motion, seeking an equilibrium price despite the inherent randomness in financial markets.

\subsubsection{Analysis based on SDEs}

Both diffusion models and the Black-Scholes equation rely on \emph{stochastic differential equations (SDEs)} to describe the evolution of systems over time. This common mathematical framework allows us to draw interesting parallels between seemingly disparate domains like image generation and financial option pricing.

\paragraph{Diffusion Models}

In text-to-image diffusion models, the forward diffusion process is governed by the SDE:

\[
\mathbf{x}_t = -\frac{1}{2}\beta_t \mathbf{x}_t \, dt + \sqrt{\beta_t}\,d\mathbf{w}_t,
\]

where $\mathbf{x}_t$ is the state (i.e., the image representation) at time $t$, $\beta_t$ is a time-dependent noise coefficient, and $d\mathbf{w}_t$ denotes the increment of a Wiener process (Brownian motion).

To reconstruct the data, the model uses the reverse diffusion process:

\[
\mathbf{x}_t = \left( \frac{1}{2}\beta_t \mathbf{x}_t + \nabla_{\mathbf{x}_t}\log p_t(\mathbf{x}_t) \right)dt + \sqrt{\beta_t}\,d\mathbf{w}_t,
\]

where $\nabla_{\mathbf{x}_t}\log p_t(\mathbf{x}_t)$ is the score function, guiding the reverse process.

\paragraph{Black-Scholes Equation}

In finance, the Black-Scholes framework models the stock price $S_t$ as following a geometric Brownian motion:

\[
dS_t = \mu S_t\,dt + \sigma S_t\,dW_t,
\]

with $\mu$ as the drift rate, $\sigma$ as the volatility, and $dW_t$ as the Wiener process. 

Using It\^o's Lemma, the evolution of the option price $V(S,t)$ is given by:

\[
dV = \left( \frac{\partial V}{\partial t} + \mu S \frac{\partial V}{\partial S} + \frac{1}{2}\sigma^2 S^2 \frac{\partial^2 V}{\partial S^2} \right) dt + \sigma S\frac{\partial V}{\partial S}\,dW_t.
\]

This derivation leads to the well-known Black-Scholes partial differential equation (PDE):

\[
\frac{\partial V}{\partial t} + rS\frac{\partial V}{\partial S} + \frac{1}{2}\sigma^2 S^2 \frac{\partial^2 V}{\partial S^2} - rV = 0,
\]

where $r$ is the risk-free interest rate.

\paragraph{Connecting the Two Domains}

Both approaches yield PDEs describing the evolution of key quantities:
\begin{enumerate}
  \item In diffusion models, the Fokker-Planck equation governs the probability density $p(\mathbf{x},t)$ of the image states:

\[
  \frac{\partial p(\mathbf{x},t)}{\partial t} = -\nabla \cdot \left( \mathbf{f}(\mathbf{x},t)p(\mathbf{x},t) \right) + \frac{1}{2}\nabla \cdot \left( D(\mathbf{x},t)\nabla p(\mathbf{x},t) \right),
  \]

  where $\mathbf{f}(\mathbf{x},t)$ is the drift term and $D(\mathbf{x},t)$ the diffusion coefficient.
  
  \item In the Black-Scholes model, the option price $V(S,t)$ evolves according to a PDE derived from the stock's stochastic dynamics.
\end{enumerate}

Moreover, both models use gradients to guide their respective processes. In diffusion models, the score function $\nabla_{\mathbf{x}_t}\log p_t(\mathbf{x}_t)$ directs the reverse diffusion process, while in the Black-Scholes framework, the gradient of $V(S,t)$ with respect to $S$ (often referred to as delta) is central to constructing a risk-neutral portfolio.

This similarity highlights a fascinating cross-domain synergy where concepts from financial mathematics can inspire novel techniques in generative modeling, and vice versa.

\begin{figure*}[h]
    \centering
    \includegraphics[width=0.99\textwidth]{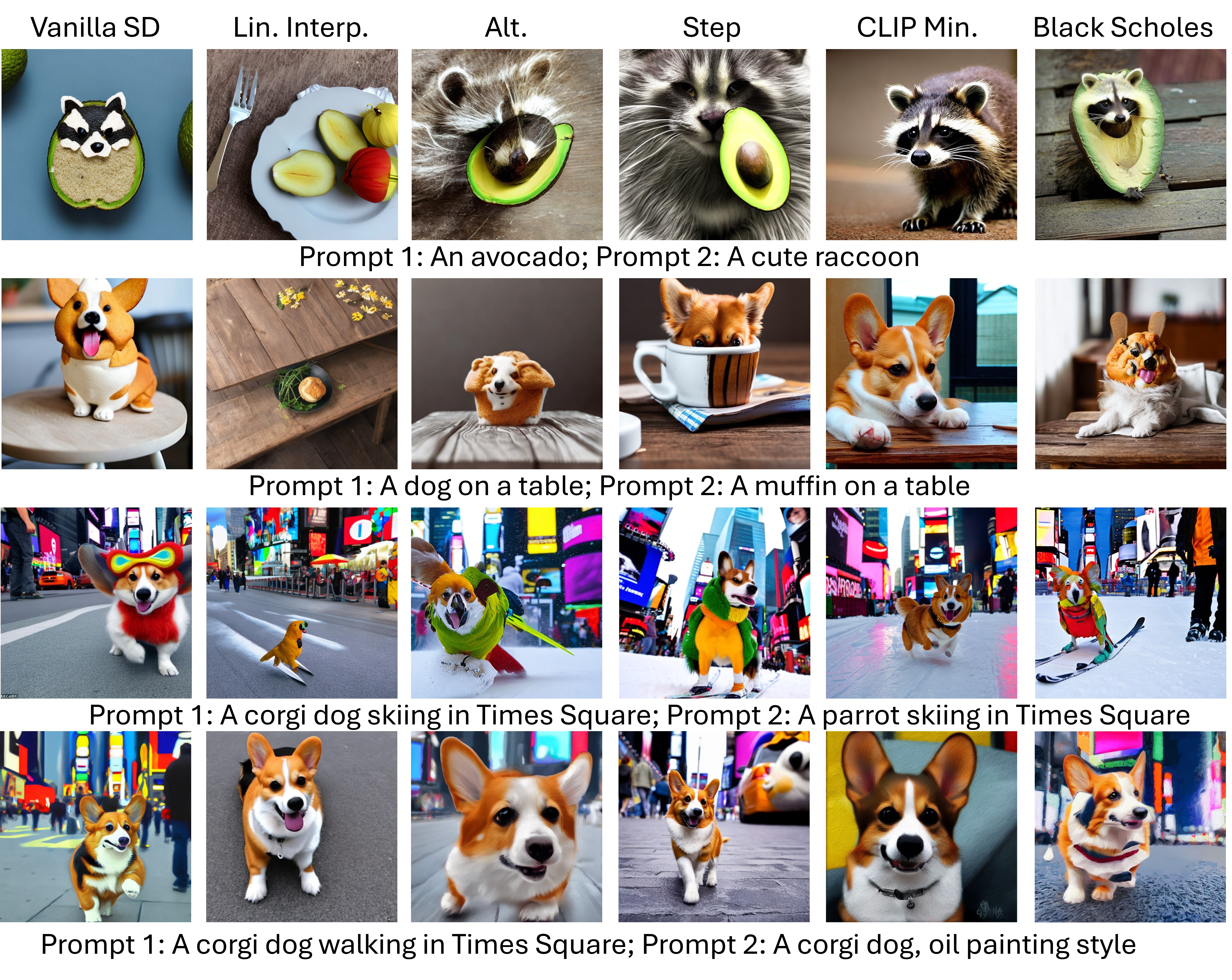}
    \vspace*{-1.25em}
    \caption{We present more results of our method along with comparisons. Vanilla SD fails to capture clear characteristics of individual text prompts, omitting distinct features such as those related to avocado/raccoon, muffin, parrot, and oil painting style. Linear interpolation generates images not consistent with the prompts, due to issues with non-linear manifolds. CLIP Min. generates images biased towards one of the prompts. Alt. and Step prompt selection methods suffer from artifacts and are not very successful in blending the characteristics of objects corresponding to the individual text prompts - the avocado/raccoon, muffin/dog are not blended well. In the parrot/dog image, the characteristics of the parrot are missing. Alt. generates artifacts in the Times Square/ oil painting image, while Times Square is not characterized well in the image generated by Step.  In contrast, the Black-Scholes model adeptly overcomes these limitations, generating realistic images that meticulously balance and preserve the unique characteristics of each individual text prompt. The images are from set 1, set 2, set 3 and set 4 (Refer Section~\ref{sec:data}) respectively.
    }
    \label{fig:main2}
    \vspace*{-1em}
\end{figure*}

\section{Blending Concepts using the Black Scholes Algorithm}

\subsection{Problem Formulation} Consider a set of $N$ text prompts denoted as $P_{1}, ... P_{N}$ and $T$ diffusion denoising steps. Note that text-to-image diffusion architectures often include a text encoder that maps the text prompt to a joint text-image space. Starting from random noise and conditioned on the embedding in the text-image space, the neural network generates the final image.

Our objective is to generate an image that aligns with multiple text prompts simultaneously i.e. at the intersection of the various text-image manifolds of the diffusion models. We pose this task as a prompt mixing or concept blending~\cite{patashnik2023localizing} problem. During each step of diffusion denoising, our aim is to select the most relevant text prompt (or the text prompt with respect to which the image requires further refining) for conditioning the model, ensuring the generation of an optimal image satisfying all individual text prompts. 

\subsection{Method}

Selecting the most effective text prompt to condition a text-to-image diffusion model can be approached in various ways. Naively switching~\cite{kothandaraman2023aerial,patashnik2023localizing} between prompts is sub-optimal—not only does it require significant human intervention, but it also fails to automatically prioritize the most relevant prompt. An alternative approach computes the CLIP score at each step and moves toward the prompt with the lowest CLIP score; however, this method overlooks the dynamic aspects inherent in the diffusion denoising process~\cite{ho2020denoising}. To address these issues, we leverage the Black-Scholes model to effectively capture these dynamics.

\subsubsection{Black-Scholes Analogy in Diffusion Models}

Our approach uses the CLIP score to quantify the ``price'' of the generated image, treating it as the underlying asset in a stock pricing analogy. The CLIP score—which is widely used to assess text-image alignment—provides a natural metric for how well the image aligns with each text prompt. For diffusion denoising at step \( i \) out of a total of \( T \) steps, we derive the Black-Scholes variables as follows:

\begin{enumerate}
    \item \textbf{Underlying Stock Price, \( S \):} Analogous to the current value of an asset, we define \( S \) as the CLIP score measuring the alignment of the generated image with a specific text prompt \( P_p \) at the current stage. Let \( z_{t} \) denote the predicted latents at timestep \( t \), and \( z_{0,t} \) represent the latents of the final predicted image extrapolated from \( z_{t} \). The score is scaled (multiplied by 100) to lie within the range \(0\) to \(100\).
    
    \item \textbf{Strike Price, \( K \):} This represents the asset's price at expiration. In our formulation, \( K \) is defined as the average CLIP score observed at the final diffusion step when the model generates an image using a straightforward combination of prompts. This value reflects the model’s maximum potential for that set of prompts.
    
    \item \textbf{Time to Expiration, \( t \):} In diffusion models, \( t \) corresponds to the number of remaining denoising steps, i.e., \( t = T - i \), where \( i \) is the current step.
    
    \item \textbf{Volatility, \( \sigma \):} In finance, volatility is the standard deviation of the asset price fluctuations. Analogously, we compute \( \sigma \) as the square root of the variance used by the diffusion denoising scheduler at timestep \( i \), reflecting the uncertainty inherent in the diffusion process.
    
    \item \textbf{Risk-free Rate, \( r \):} In traditional finance, the risk-free rate is determined by economic factors. For our purposes, we simplify by setting \( r = \frac{1}{T} \) to distribute returns equally over the entire diffusion process.
\end{enumerate}

\subsubsection{Black-Scholes Guided Prompt Selection}

At each iteration \( i \) of the diffusion denoising process, we compute the Black-Scholes score \( b_{i,p} \) for each text prompt \( P_p \) (with \( p = 1, \dots, N \)) using the variables introduced above along with Equations~\ref{eq:bs} and~\ref{eq:bs_d}. In the subsequent diffusion step, the model conditions on the text prompt associated with the lowest Black-Scholes score. This ensures that the denoising process prioritizes the prompt that most needs refinement, enhancing the overall alignment of the generated image with all inputs.

For a detailed step-by-step description of our method, please refer to Algorithm~\ref{algbox:algorithm}.

\begin{algorithm*}
\caption{Black Scholes concept blending in backward diffusion enables the diffusion model generate an optimal image at the intersection of multiple text-image manifolds.}
\label{algbox:algorithm}
\begin{algorithmic}[1]
\State Initialize latents to random Gaussian noise. $z_T\sim\mathcal N(0,I)$; T is the total number of diffusion denoising steps. 
\State Use the text encoder $\epsilon$ to compute the CLIP embeddings of a (linguistic) combination of the text prompts $\{P\}$. $e \gets \epsilon_{p}(\cup \{P\})$.
\State Initialize Black Scholes variables, strike price $K = 0.25 \times 100$, $rate r = 1/T$.
\State // \textit{Diffusion denoising - image prediction/generation loop}
\For{$t\gets T$ to $0$}
    \State $z_{t-1}=z_t - f(z_t, t, e)$; f is the diffusion UNet.
    \State Use $z_{t-1}$ to compute $z_{0,t-1}$.
    \State Variance $\sigma$ at step $t-1$ $\gets$ computed using the scheduler of the diffusion model. 
    \For{$i\gets 1$ to $N$}
        \State Spot price $S$ $\gets$ CLIP score With respect to text prompt $P_{i}$
        \State Time to expiration $\gets t$
        \State Black Scholes score $b_{t,i}$ with respect to prompt $P_{i}$, at timestep $t$, $\gets$ use Equation 1,2. 
    \EndFor
    \State $P_{min}$ $\gets$ Text prompt corresponding to min$\{B_{t,i}\}$, $i=1...N$
    \State $e$ $\gets$ $\epsilon_{p}(P_{min})$
\EndFor
\end{algorithmic}
\end{algorithm*}

\section{Experiments and Results}

\paragraph{Metrics.}

We evaluate performance using the following metrics:

\begin{enumerate}
    \item \textbf{Text-image alignment}: We utilize two variants of the CLIP score:
    \begin{itemize}
        \item \textbf{CLIP-combined}: This variant assesses the overall alignment with text by comparing the generated image against a combination of individual text prompts.
        \item \textbf{CLIP-add}: This averages the CLIP scores for the generated image across each individual text prompt, reflecting alignment with each specific concept.
    \end{itemize}
    \item \textbf{Preserving fidelity (or characteristics) w.r.t. each constituent concept}:
    \begin{itemize}
        \item The \textbf{BLIP score} is calculated by comparing the generated image to a combination of individual text prompts, measuring overall text-level alignment.
        \item The \textbf{DINO score} evaluates the generated image against each individual text prompt, assessing how well the image preserves the characteristics of each concept. This indicates the fidelity of the attributes in the generated image relative to the individual concepts. 
    \end{itemize}
    Thus, while BLIP focuses on overall concept blending, DINO provides insights into the preservation of characteristics for each concept. We aim for high values in both scores, calculating a net score by multiplying the DINO and BLIP scores for each prompt and averaging across all prompts.
    \item \textbf{Generation quality}: We use KID to assess the realism of the generated samples, serving as an indicator of their overall quality.
\end{enumerate}

To calculate our metrics, we generate five images for each text prompt and incorporate all of them into the metric computations.

\paragraph{Baselines.} Our study examines several baselines: (i) \textbf{Vanilla SD:} Prompt engineering, we use the vanilla stable diffusion model and condition it on a single text prompt effectively describing all individual text prompts, (ii) \textbf{Linear Interpolation:} Direct combination of text embeddings, achieved through linear interpolation between text embeddings~\cite{kawar2023imagic}. This method equally weights the text embeddings associated with each text prompt. (iii) \textbf{Switching between text prompts}. We consider two variations here: (iii-a) \textbf{Step:} Following Patashnik et. al.~\cite{patashnik2023localizing}, we use one text prompt for the initial 7th to 17th denoising steps and then switch to the other text prompt for the remaining steps, (iii-b) \textbf{Alt.:} Following Kothandaraman et. al.~\cite{kothandaraman2023aerial}, we alternate between the two text prompts. (iv) \textbf{CLIP Minimum:} Score-based combination, denoted as CLIP-min, where we select the text prompt corresponding to the lowest CLIP score from the previous denoising iteration.

\paragraph{Backbone architecture.} We use the Stable Diffusion 2.1 backbone model~\cite{rombach2022high} in all our experiments. There is no training involved, we use the pretrained model to directly perform inference. Our experiments are run on one NVIDIA A5000 GPU with 24 GB memory.

\begin{table*}
    \centering
    \resizebox{0.99\textwidth}{!}{
    \begin{tabular}{ccccccccc}
    \toprule 
        Method & CLIP-combined ($\uparrow$) & CLIP-add ($\uparrow$) & BLIP $\odot$ DINO ($\uparrow$) & KID ($\downarrow$) & Steps ($\downarrow$) & Time (s) ($\downarrow$) & GPU hrs & Memory (GB) ($\downarrow$)\\
        \midrule 
        Linear Int.~\cite{kawar2023imagic} & $0.2885$ & $0.2778$ & $0.2588$ & $0.02851$ & $50$ & $6.5$ & $0.001805$ & $7.1$\\
        Alt. Samp.~\cite{kothandaraman2023aerial} & $0.3445$ & $0.3098$ & $0.3894$ & $0.01786$ & $100$ & $14$ & $0.00389$ & $7.7$ \\
        CLIP Min. & $0.3195$ & $0.2955$ & $0.3107$ & $0.00866$ & $100$ & $14$ & $0.00389$ & $7.7$\\
        Step~\cite{patashnik2023localizing} & $0.3220$ & $0.2997$ & $0.3390$ & $0.01709$ & $100$ & $14$ & $0.00389$ & $7.7$ \\
        Black Scholes & {\bf 0.3469} & {\bf 0.3112} & {\bf 0.3912} & {\bf 0.01531} & {\bf 100} & {\bf 14} & {\bf 0.00389} & {\bf 7.7} \\
        
       \bottomrule
    \end{tabular}
    }
    \caption{We evaluate various properties using CLIP Scores, BLIP Scores, DINO, and KID. These metrics help us assess overall text alignment with the combined text prompts, the preservation of attributes related to individual concepts, and the quality of the generated images. The Black Scholes algorithm for concept blending in diffusion models achieves superior results, compared to other prompt-mixing techniques, as also evidenced by the qualitative results.}
    \label{tab:clip}
\end{table*}

\paragraph{Hyperparameters.} Based on our experiments for the vanilla combination using Stable Diffusion 2.1 for the dataset under consideration, where we found that a CLIP score of approximately 0.25 indicates reasonable text-image alignment, we opted for a constant value of 0.25 for the strike price. The ordering of prompts does not matter. This is because, at every step of diffusion denoising, the algorithm chooses the prompt that should be selected by computing the Black Scholes score wrt each prompt, and this process is agnostic to the ordering of the prompts. More details on hyperparameters and the backbone architecture can be found in the appendix. 



\begin{figure}
    \centering
    \includegraphics[width=0.5\textwidth]{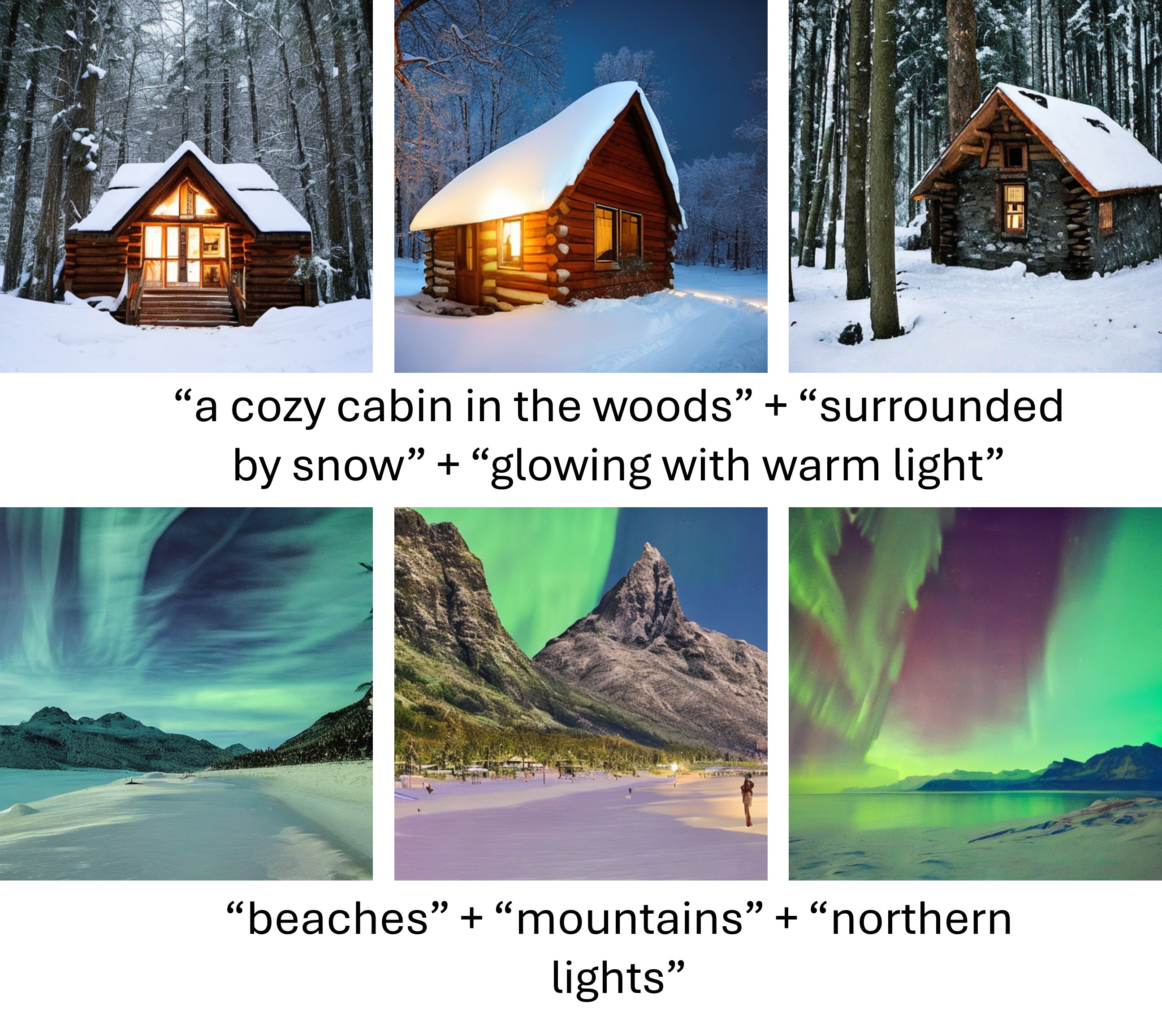}
    \vspace{-1em}
    \caption{Our method can be extended to concept blending involving more than 2 prompts (3 in this case), as shown above.}
    \label{fig:supp1}
\end{figure}

\subsection{Analysis and Comparisons}

\label{sec:data}
We construct a dataset with $4$ types of scenarios to analyze varying complexities, spanning:
\begin{itemize}
    \item \textbf{Simple text prompts:} A single object per prompt.
    \item \textbf{Multiple objects:} Multiple objects described within one prompt.
    \item \textbf{Object actions against backgrounds:} Concept blending with respect to the object.
    \item \textbf{Object performing action against background/style:} Concept blending with respect to background and style.
\end{itemize}
More details can be found in the appendix.

Qualitative results are shown in Figure~\ref{fig:main1} and Figure~\ref{fig:main2}, with additional examples provided in the appendix. Quantitative comparisons are summarized in Figure~\ref{tab:clip}. A detailed analysis of our method relative to prior art is described below:

\begin{itemize}
    \item \textbf{Vanilla Stable Diffusion (SD):} Vanilla Stable Diffusion v2.1 can generate plausible combinations of the input text prompts. However, due to \textit{hallucination issues} and constraints imposed by the training distribution, it struggles to capture the distinct characteristics of individual prompts. For instance, in Figure~\ref{fig:main1} the unique features of the parrot, cat, dog/penguin, and sunset/penguin are not well preserved, and similarly, Figure~\ref{fig:main2} shows a lack of distinct characteristics for avocado/raccoon, muffin, parrot, and the oil painting style.
    
    \item \textbf{Linear Interpolation~\cite{kawar2023imagic}:} Linear interpolation, which assumes a simple linear relationship, is ill-suited for blending two text prompts due to the highly non-linear nature of the text-image space. It only traverses a straight path between endpoints in the latent space, failing to capture nuanced variations or generate novel features.
    
    \item \textbf{Alternating Sampling~\cite{kothandaraman2023aerial}:} While alternating sampling produces features related to both prompts in the final output image, it often results in poorly defined objects, unrealistic appearance, and low overall quality. Artifacts are prevalent, and many images appear implausible, leading to a high KID score. The main issue is that the method alternates between the two text prompts without emphasizing the most relevant one during each denoising step.
    
    \item \textbf{Step-wise Switching~\cite{kothandaraman2023aerial}:} This technique suffers from similar issues as alternating sampling and fails to effectively capture the detailed characteristics of multiple text prompts.
    
    \item \textbf{CLIP-min:} The CLIP-min approach tends to favor one text prompt over the other, resulting in images that do not adequately reflect the intersection of both prompts. It selects the text prompt with the lowest CLIP score at each step, thereby overlooking the dynamic nature of the diffusion denoising process.
    
    \item \textbf{Black-Scholes Algorithm:} Our proposed method, based on the Black-Scholes algorithm, effectively generates realistic images that align with the intersection of the two text prompts. Our approach \textit{preserves the unique attributes} of each prompt while minimizing unrealistic artifacts. By modeling the dynamics of the diffusion denoising process, our algorithm strategically selects the optimal text prompt at every step, achieving a balanced synthesis and superior quantitative performance.
\end{itemize}

In summary, our Black-Scholes model–inspired method outperforms previous prompt mixing approaches by generating realistic images that meticulously preserve and balance the features of each individual text prompt.

\paragraph{Computational complexity}

We report the computational requirements for different methods in Table~\ref{tab:clip}. All values are computed using one NVIDIA A5000 GPU, on the 2 prompts blending scenario. Our algorithm incurs minimal computational overhead over other methods, primarily arising from the computation of the CLIP score of the generated image w.r.t. all individual text prompts at every step of the diffusion denoising based generation process. 
\section{Conclusions, Limitations, and Future Work}

In this paper, we introduced a novel approach to concept blending for text-to-image diffusion models that is deeply inspired by financial probabilistic models, particularly the Black-Scholes framework. By drawing an analogy between asset pricing and the dynamics of the diffusion process, we developed an automated algorithm that dynamically selects the most relevant text prompt at each denoising step. This approach not only enhances the alignment between the generated image and the input prompts but also offers a data-efficient alternative to conventional prompt mixing methods.

Our extensive qualitative and quantitative evaluations demonstrate that our method outperforms traditional techniques such as linear interpolation, alternating sampling, and manual prompt switching. While our approach leverages the strengths of modern diffusion models, it is important to note that it may not extend to non-Gaussian diffusion processes~\cite{bansal2024cold} or one-step diffusion models~\cite{yin2023one}—a limitation that opens interesting avenues for future research.

Moreover, the underlying principles of our method have significant implications beyond concept blending. They can be readily integrated into state-of-the-art text-to-image architectures, and they pave the way for critical downstream tasks including image editing~\cite{kawar2023imagic,yang2023paint,avrahami2022blended}, compositional synthesis~\cite{liu2022compositional,agarwal2023star}, handling complex prompts~\cite{lian2023llm}, text-based view synthesis~\cite{kothandaraman2023aerial,kothandaraman2023aerialbooth}, and personalized image generation~\cite{ruiz2023dreambooth,kumari2023multi}. Our work demonstrates that financially inspired methods have the potential to significantly enhance concept blending in text-to-image generation, thereby opening avenues for broader innovation across the field of generative AI.

\paragraph{Impact Statement.} Our approach advances the state-of-the-art in generative AI for content creation by harnessing the synergy between diffusion models and financial probabilistic frameworks. However, the increased realism and flexibility afforded by our method also raise significant concerns regarding potential misuse. This dual-edged potential underscores the urgent need for continued research in watermarking and deepfake detection to responsibly mitigate associated risks while enabling creative innovations.

\paragraph{Acknowledgements.} This work was supported in part by ARO Grants W911NF2110026, W911NF2310046,  W911NF2310352  and Army Cooperative Agreement W911NF2120076.

{\small
\bibliographystyle{ref_style}
\bibliography{references}

\begin{thebibliography}{10}\itemsep=-1pt

\bibitem{abdal2019image2stylegan}
Rameen Abdal, Yipeng Qin, and Peter Wonka.
\newblock Image2stylegan: How to embed images into the stylegan latent space?
\newblock In {\em Proceedings of the IEEE/CVF international conference on computer vision}, pages 4432--4441, 2019.

\bibitem{adobefirefly}
Adobe.
\newblock Adobe firefly.
\newblock {\em https://firefly.adobe.com/}, 2024.

\bibitem{agarwal2023star}
Aishwarya Agarwal, Srikrishna Karanam, KJ Joseph, Apoorv Saxena, Koustava Goswami, and Balaji~Vasan Srinivasan.
\newblock A-star: Test-time attention segregation and retention for text-to-image synthesis.
\newblock In {\em Proceedings of the IEEE/CVF International Conference on Computer Vision}, pages 2283--2293, 2023.

\bibitem{avrahami2022blended}
Omri Avrahami, Dani Lischinski, and Ohad Fried.
\newblock Blended diffusion for text-driven editing of natural images.
\newblock In {\em Proceedings of the IEEE/CVF Conference on Computer Vision and Pattern Recognition}, pages 18208--18218, 2022.

\bibitem{balaji2022ediff}
Yogesh Balaji, Seungjun Nah, Xun Huang, Arash Vahdat, Jiaming Song, Qinsheng Zhang, Karsten Kreis, Miika Aittala, Timo Aila, Samuli Laine, et~al.
\newblock ediff-i: Text-to-image diffusion models with an ensemble of expert denoisers.
\newblock {\em arXiv preprint arXiv:2211.01324}, 2022.

\bibitem{bansal2024cold}
Arpit Bansal, Eitan Borgnia, Hong-Min Chu, Jie Li, Hamid Kazemi, Furong Huang, Micah Goldblum, Jonas Geiping, and Tom Goldstein.
\newblock Cold diffusion: Inverting arbitrary image transforms without noise.
\newblock {\em Advances in Neural Information Processing Systems}, 36, 2024.

\bibitem{betker2023improving}
James Betker, Gabriel Goh, Li Jing, Tim Brooks, Jianfeng Wang, Linjie Li, Long Ouyang, Juntang Zhuang, Joyce Lee, Yufei Guo, et~al.
\newblock Improving image generation with better captions.
\newblock {\em Computer Science. https://cdn. openai. com/papers/dall-e-3. pdf}, 2(3):8, 2023.

\bibitem{black2019pricing}
Fischer Black and Myron Scholes.
\newblock The pricing of options and corporate liabilities.
\newblock In {\em World Scientific Reference on Contingent Claims Analysis in Corporate Finance: Volume 1: Foundations of CCA and Equity Valuation}, pages 3--21. World Scientific, 2019.

\bibitem{chefer2023attend}
Hila Chefer, Yuval Alaluf, Yael Vinker, Lior Wolf, and Daniel Cohen-Or.
\newblock Attend-and-excite: Attention-based semantic guidance for text-to-image diffusion models.
\newblock {\em ACM Transactions on Graphics (TOG)}, 42(4):1--10, 2023.

\bibitem{cheng2023layoutdiffuse}
Jiaxin Cheng, Xiao Liang, Xingjian Shi, Tong He, Tianjun Xiao, and Mu Li.
\newblock Layoutdiffuse: Adapting foundational diffusion models for layout-to-image generation.
\newblock {\em arXiv preprint arXiv:2302.08908}, 2023.

\bibitem{esser2024scaling}
Patrick Esser, Sumith Kulal, Andreas Blattmann, Rahim Entezari, Jonas M{\"u}ller, Harry Saini, Yam Levi, Dominik Lorenz, Axel Sauer, Frederic Boesel, et~al.
\newblock Scaling rectified flow transformers for high-resolution image synthesis.
\newblock {\em arXiv preprint arXiv:2403.03206}, 2024.

\bibitem{gal2022stylegan}
Rinon Gal, Or Patashnik, Haggai Maron, Amit~H Bermano, Gal Chechik, and Daniel Cohen-Or.
\newblock Stylegan-nada: Clip-guided domain adaptation of image generators.
\newblock {\em ACM Transactions on Graphics (TOG)}, 41(4):1--13, 2022.

\bibitem{ho2020denoising}
Jonathan Ho, Ajay Jain, and Pieter Abbeel.
\newblock Denoising diffusion probabilistic models.
\newblock {\em Advances in neural information processing systems}, 33:6840--6851, 2020.

\bibitem{ho2022cascaded}
Jonathan Ho, Chitwan Saharia, William Chan, David~J Fleet, Mohammad Norouzi, and Tim Salimans.
\newblock Cascaded diffusion models for high fidelity image generation.
\newblock {\em Journal of Machine Learning Research}, 23(47):1--33, 2022.

\bibitem{hong2023improving}
Susung Hong, Gyuseong Lee, Wooseok Jang, and Seungryong Kim.
\newblock Improving sample quality of diffusion models using self-attention guidance.
\newblock In {\em Proceedings of the IEEE/CVF International Conference on Computer Vision}, pages 7462--7471, 2023.

\bibitem{karras2017progressive}
Tero Karras, Timo Aila, Samuli Laine, and Jaakko Lehtinen.
\newblock Progressive growing of gans for improved quality, stability, and variation.
\newblock {\em arXiv preprint arXiv:1710.10196}, 2017.

\bibitem{kawar2023imagic}
Bahjat Kawar, Shiran Zada, Oran Lang, Omer Tov, Huiwen Chang, Tali Dekel, Inbar Mosseri, and Michal Irani.
\newblock Imagic: Text-based real image editing with diffusion models.
\newblock In {\em Proceedings of the IEEE/CVF Conference on Computer Vision and Pattern Recognition}, pages 6007--6017, 2023.

\bibitem{kothandaraman2023aerial}
Divya Kothandaraman, Tianyi Zhou, Ming Lin, and Dinesh Manocha.
\newblock Aerial diffusion: Text guided ground-to-aerial view translation from a single image using diffusion models.
\newblock {\em arXiv preprint arXiv:2303.11444}, 2023.

\bibitem{kothandaraman2023aerialbooth}
Divya Kothandaraman, Tianyi Zhou, Ming Lin, and Dinesh Manocha.
\newblock Aerialbooth: Mutual information guidance for text controlled aerial view synthesis from a single image.
\newblock {\em arXiv preprint arXiv:2311.15478}, 2023.

\bibitem{kumari2023multi}
Nupur Kumari, Bingliang Zhang, Richard Zhang, Eli Shechtman, and Jun-Yan Zhu.
\newblock Multi-concept customization of text-to-image diffusion.
\newblock In {\em Proceedings of the IEEE/CVF Conference on Computer Vision and Pattern Recognition}, pages 1931--1941, 2023.

\bibitem{kwon2022diffusion}
Mingi Kwon, Jaeseok Jeong, and Youngjung Uh.
\newblock Diffusion models already have a semantic latent space.
\newblock {\em arXiv preprint arXiv:2210.10960}, 2022.

\bibitem{levin2017markov}
David~A Levin and Yuval Peres.
\newblock {\em Markov chains and mixing times}, volume 107.
\newblock American Mathematical Soc., 2017.

\bibitem{lian2023llm}
Long Lian, Boyi Li, Adam Yala, and Trevor Darrell.
\newblock Llm-grounded diffusion: Enhancing prompt understanding of text-to-image diffusion models with large language models.
\newblock {\em arXiv preprint arXiv:2305.13655}, 2023.

\bibitem{liu2022compositional}
Nan Liu, Shuang Li, Yilun Du, Antonio Torralba, and Joshua~B Tenenbaum.
\newblock Compositional visual generation with composable diffusion models.
\newblock In {\em European Conference on Computer Vision}, pages 423--439. Springer, 2022.

\bibitem{merton1976option}
Robert~C Merton.
\newblock Option pricing when underlying stock returns are discontinuous.
\newblock {\em Journal of financial economics}, 3(1-2):125--144, 1976.

\bibitem{patashnik2023localizing}
Or Patashnik, Daniel Garibi, Idan Azuri, Hadar Averbuch-Elor, and Daniel Cohen-Or.
\newblock Localizing object-level shape variations with text-to-image diffusion models.
\newblock In {\em Proceedings of the IEEE/CVF International Conference on Computer Vision}, pages 23051--23061, 2023.

\bibitem{imagemixerdiffusion}
Justin Pinkney.
\newblock Image mixer diffusion.
\newblock {\em https://www.justinpinkney.com/blog/2024/image-mixer-diffusion/}, 2024.

\bibitem{preechakul2022diffusion}
Konpat Preechakul, Nattanat Chatthee, Suttisak Wizadwongsa, and Supasorn Suwajanakorn.
\newblock Diffusion autoencoders: Toward a meaningful and decodable representation.
\newblock In {\em Proceedings of the IEEE/CVF Conference on Computer Vision and Pattern Recognition}, pages 10619--10629, 2022.

\bibitem{CLIPGuidedImageMixing}
GitHub Repo.
\newblock Clip guided images mixing with stable diffusion.
\newblock {\em https://github.com/TheDenk/images$\_$mixing}, 2023.

\bibitem{rombach2022high}
Robin Rombach, Andreas Blattmann, Dominik Lorenz, Patrick Esser, and Bj{\"o}rn Ommer.
\newblock High-resolution image synthesis with latent diffusion models.
\newblock In {\em Proceedings of the IEEE/CVF conference on computer vision and pattern recognition}, pages 10684--10695, 2022.

\bibitem{ruiz2023dreambooth}
Nataniel Ruiz, Yuanzhen Li, Varun Jampani, Yael Pritch, Michael Rubinstein, and Kfir Aberman.
\newblock Dreambooth: Fine tuning text-to-image diffusion models for subject-driven generation.
\newblock In {\em Proceedings of the IEEE/CVF Conference on Computer Vision and Pattern Recognition}, pages 22500--22510, 2023.

\bibitem{saharia2022photorealistic}
Chitwan Saharia, William Chan, Saurabh Saxena, Lala Li, Jay Whang, Emily~L Denton, Kamyar Ghasemipour, Raphael Gontijo~Lopes, Burcu Karagol~Ayan, Tim Salimans, et~al.
\newblock Photorealistic text-to-image diffusion models with deep language understanding.
\newblock {\em Advances in neural information processing systems}, 35:36479--36494, 2022.

\bibitem{shen2020interpreting}
Yujun Shen, Jinjin Gu, Xiaoou Tang, and Bolei Zhou.
\newblock Interpreting the latent space of gans for semantic face editing.
\newblock In {\em Proceedings of the IEEE/CVF conference on computer vision and pattern recognition}, pages 9243--9252, 2020.

\bibitem{sohl2015deep}
Jascha Sohl-Dickstein, Eric Weiss, Niru Maheswaranathan, and Surya Ganguli.
\newblock Deep unsupervised learning using nonequilibrium thermodynamics.
\newblock In {\em International conference on machine learning}, pages 2256--2265. PMLR, 2015.

\bibitem{witteveen2022investigating}
Sam Witteveen and Martin Andrews.
\newblock Investigating prompt engineering in diffusion models.
\newblock {\em arXiv preprint arXiv:2211.15462}, 2022.

\bibitem{wu2023self}
Tsung-Han Wu, Long Lian, Joseph~E Gonzalez, Boyi Li, and Trevor Darrell.
\newblock Self-correcting llm-controlled diffusion models.
\newblock {\em arXiv preprint arXiv:2311.16090}, 2023.

\bibitem{yang2023paint}
Binxin Yang, Shuyang Gu, Bo Zhang, Ting Zhang, Xuejin Chen, Xiaoyan Sun, Dong Chen, and Fang Wen.
\newblock Paint by example: Exemplar-based image editing with diffusion models.
\newblock In {\em Proceedings of the IEEE/CVF Conference on Computer Vision and Pattern Recognition}, pages 18381--18391, 2023.

\bibitem{yang2023diffusion}
Yongqi Yang, Ruoyu Wang, Zhihao Qian, Ye Zhu, and Yu Wu.
\newblock Diffusion in diffusion: Cyclic one-way diffusion for text-vision-conditioned generation.
\newblock {\em arXiv preprint arXiv:2306.08247}, 2023.

\bibitem{yin2023one}
Tianwei Yin, Micha{\"e}l Gharbi, Richard Zhang, Eli Shechtman, Fredo Durand, William~T Freeman, and Taesung Park.
\newblock One-step diffusion with distribution matching distillation.
\newblock {\em arXiv preprint arXiv:2311.18828}, 2023.

\bibitem{zheng2023layoutdiffusion}
Guangcong Zheng, Xianpan Zhou, Xuewei Li, Zhongang Qi, Ying Shan, and Xi Li.
\newblock Layoutdiffusion: Controllable diffusion model for layout-to-image generation.
\newblock In {\em Proceedings of the IEEE/CVF Conference on Computer Vision and Pattern Recognition}, pages 22490--22499, 2023.

\bibitem{zhu2020domain}
Jiapeng Zhu, Yujun Shen, Deli Zhao, and Bolei Zhou.
\newblock In-domain gan inversion for real image editing.
\newblock In {\em European conference on computer vision}, pages 592--608. Springer, 2020.

\bibitem{zhu2016generative}
Jun-Yan Zhu, Philipp Kr{\"a}henb{\"u}hl, Eli Shechtman, and Alexei~A Efros.
\newblock Generative visual manipulation on the natural image manifold.
\newblock In {\em Computer Vision--ECCV 2016: 14th European Conference, Amsterdam, The Netherlands, October 11-14, 2016, Proceedings, Part V 14}, pages 597--613. Springer, 2016.

\bibitem{zhu2023unseen}
Ye Zhu, Yu Wu, Zhiwei Deng, Olga Russakovsky, and Yan Yan.
\newblock Unseen image synthesis with diffusion models.
\newblock {\em arXiv preprint arXiv:2310.09213}, 2023.

\bibitem{zhu2024boundary}
Ye Zhu, Yu Wu, Zhiwei Deng, Olga Russakovsky, and Yan Yan.
\newblock Boundary guided learning-free semantic control with diffusion models.
\newblock {\em Advances in Neural Information Processing Systems}, 36, 2024.

\end{thebibliography}
}

\section*{A.1. Experimental Settings}

We consider the following experimental settings: 

\paragraph{Set 1:} Simple text prompts with single objects. In this experiment, we validate our approach using straightforward scenarios, we use two text prompts, each describing one class. Specifically, we consider $17$ classes: [`a rock', `a coffee mug', `a cute dog', `a pink cat', `a teddy bear', `a robot', `an alien', `an avocado', `a cute raccoon', `a corgi dog', `a parrot', `a car', `a squirrel', `a cute rabbit', `pizza', `muffin', `icecream'], and construct $17c_{2} = 136$ text prompts.
\paragraph{Set 2:} Multiple objects per text prompt. To add complexity, blend multiple objects, in the presence of additional objects in the scene. The goal of this experiment is to assess the models' capabilities in blending the right objects in the scene, and preserving the characteristics of the other objects. We work with three classes of objects: [`a basket', `a teapot'], [`apples', `bananas', `a cute cat', `a cute dog', `muffins'], [`a table', `a carpet', `a bed'] and construct combinations using the first and second set and the second and third set. Additionally, we include the data point “a cat in a bathtub” and “a corgi dog in a bathtub,” resulting in a total of $71$ prompts.
\paragraph{Set 3:} Object actions against backgrounds (concept blending w.r.t. object). We investigate the model’s ability to morph objects while considering object-level action information and scene or background context. We consider the following backgrounds, [`walking in Times Square', `skiing in Times Square', `walking in a beautiful garden', `surfing on the beach', `eating watermelon on the beach', `sitting on a sofa on the beach', `watching northern lights', `watching sunset at a beach', `admiring the opera house in Sydney', `sleeping in a cozy bedroom', `admiring a beautiful waterfall in a forest', `walking in a cherry blossom garden', `walking in a colorful autumn forest', `flying in the sky at sunset']. The objects are [`a kangaroo', `a cute cat', `a corgi dog', `a parrot', `a teddy bear', `a penguin']. In total, we create $210$ prompts.
\paragraph{Set 4:} Object performing action against a background (concept blending w.r.t. background). In this experiment, we explore how the model blends global information and backgrounds while considering objects in various scenarios. We consider the objects [`a kangaroo', `a cute cat', `a corgi dog', `a parrot', `a teddy bear', `a penguin']. The backgrounds and style prompts are [`walking in Times Square', `sunset', `van gogh style', `oil painting', `a garden with tulips', `northern lights']. Overall, we have $90$ prompts in this experiment. 

\begin{figure}
    \centering
    \includegraphics[width=0.99\textwidth]{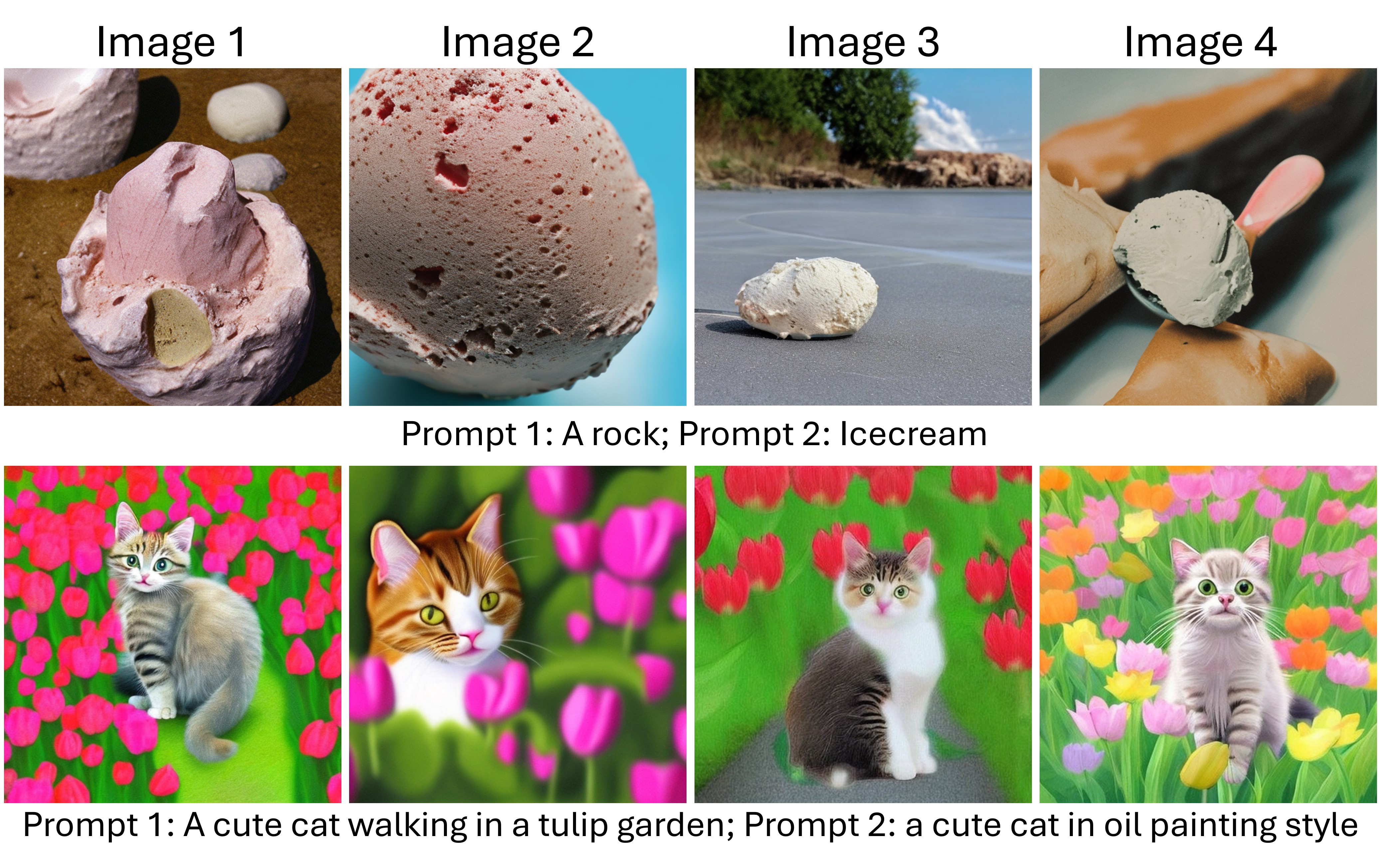}
    \vspace*{-1em}
    \caption{Image variations from our method using the Black Scholes model, starting from different random Gaussian noise initializations.}
    \label{fig:main3}
\end{figure}

\begin{figure}
    \centering
    \includegraphics[width=0.99\textwidth]{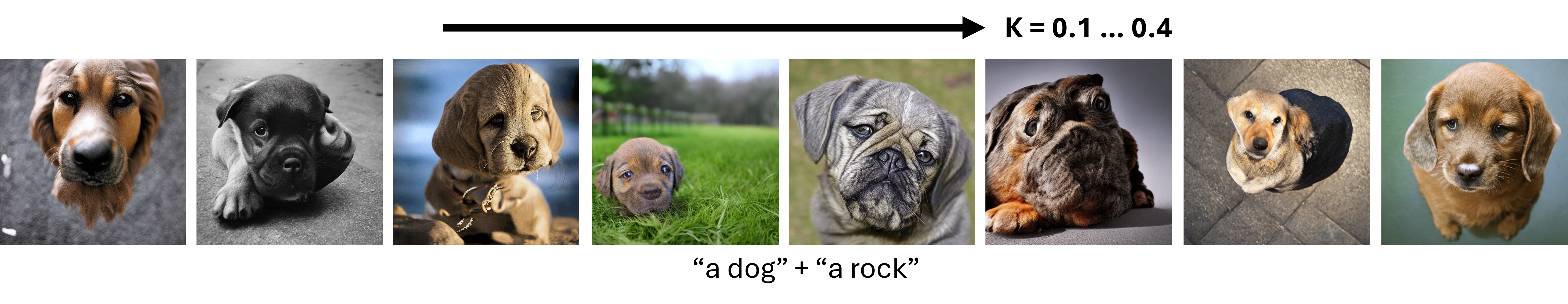}
    \caption{Ablation experiment on the strike price $K$ reveals that the results are optimal at values of $K$ close to the CLIP score of the image generated in the vanilla case using a combination of the constituent text prompts.}
    \label{fig:supp1}
\end{figure}

\subsection*{A.1.1. Hyperparameters:} We use the Stable Diffusion 2.1 backbone in all our experiments. We use standard diffusion parameters: image size - (512, 512); classifier-free guidance scale - 7.5; DDIMScheduler with betaStart set to 0.00085, betaEnd set to 0.012, betaSchedule set to ``scaledLinear'', clipSample set to False and  setAlphaToOne set to False. We use $100$ inference steps in all our experiments. 

For the set of prompts in our dataset under consideration, the average CLIP score for the combination of prompts in the vanilla case was $0.25$. Hence, we set the value of strike price $K$ at $0.25$. 

\section*{A.2. Assumptions made by the Black Scholes Algorithm}

The Black Scholes model makes five assumptions: (i) No dividends are paid out during the life of an option, (ii) Market movements are somewhat random, (iii) There are no transaction costs in buying the asset, (iv) The volatility and risk free rate of the underlying asset are known and constant, (v) The returns of the underlying asset are normally distributed. The relation between the assumptions made by the Black Scholes algorithm and diffusion models is as follows:

\begin{itemize}
    \item No Dividends (Assumption i): This assumption is specific to financial options and does not directly apply to diffusion models. In the context of diffusion models, we can consider that the "output" (the generated image) does not have any intermediate rewards or returns, similar to how options do not yield dividends.
    \item Random Market Movements (Assumption ii): This assumption aligns well with the inherent randomness in the diffusion process. Just as stock prices are subject to unpredictable fluctuations, the generation of images in diffusion models involves stochastic processes where the final output is influenced by random noise and iterative refinement.
    \item No Transaction Costs (Assumption iii): Similar to the first assumption, transaction costs are not relevant in the context of diffusion models. Instead, we can think of the computational resources and time required for generating images as analogous to transaction costs, but they do not affect the model's fundamental operation.
    \item Constant Volatility and Risk-Free Rate (Assumption iv): In diffusion models, while volatility is not explicitly defined as in financial contexts, the concept of stability in the generation process can be likened to having a consistent framework for how noise is added and refined. The ``risk-free rate'' does not have a direct counterpart, but the idea of a stable environment for generating images can be considered.
    \item Normally Distributed Returns (Assumption v): This assumption can be related to the distribution of pixel values or features in the generated images. In many diffusion models, the outputs are conditioned on a distribution that can be approximated as normal, especially when considering the latent space representations.

\end{itemize}

In summary, while some assumptions of the Black-Scholes model are not directly applicable to diffusion models, others, particularly the randomness and distribution aspects, provide a meaningful connection, which leads to our algorithm on how concepts from finance can inform generative modeling techniques!

\begin{figure}
    \centering
    \includegraphics[width=0.99\textwidth]{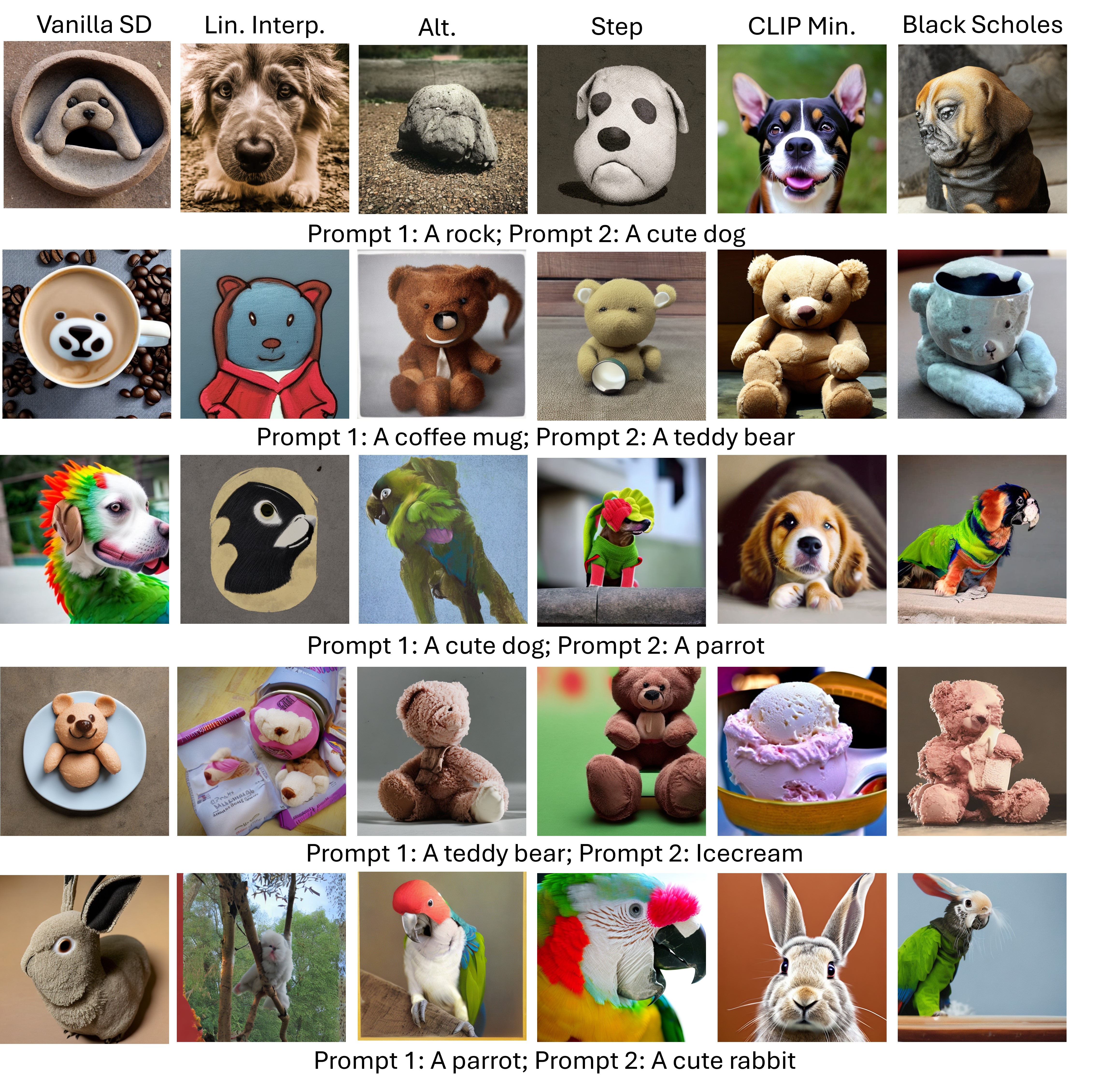}
    \caption{Vanilla stable diffusion (SD) is able to generate images satisfying both text prompts in many cases, however it misses out on the fine-grained characteristics of the individual text prompts. For instance, it misses out on the characteristics of the dog, parrot and ice-cream. Linear interpolation generates images that are not consistent with the text prompts. CLIP Min. generates images biased towards one of the text promots. Alternating sampling and Step wise inference strategies generate images with a lot of artifacts, and the generated images are perceptually implausible in many cases. For instance, in order, the issues are: missing characteristics of dog, missing characteristics of coffee mug and artifacts in teddy bear, artifacts and missing characteristics of parrot, missing characteristics of teddy bear and artifacts in teddy bear, missing characteristics of rabbit. Our Black Scholes method is able to generate images that capture the fine-grained characteristics of objects corresponding to both text prompts, and the generated images look realistic with minimal artifacts. The text prompts are from set 1.}
    \label{fig:supp1}
\end{figure}

\begin{figure}
    \centering
    \includegraphics[width=0.99\textwidth]{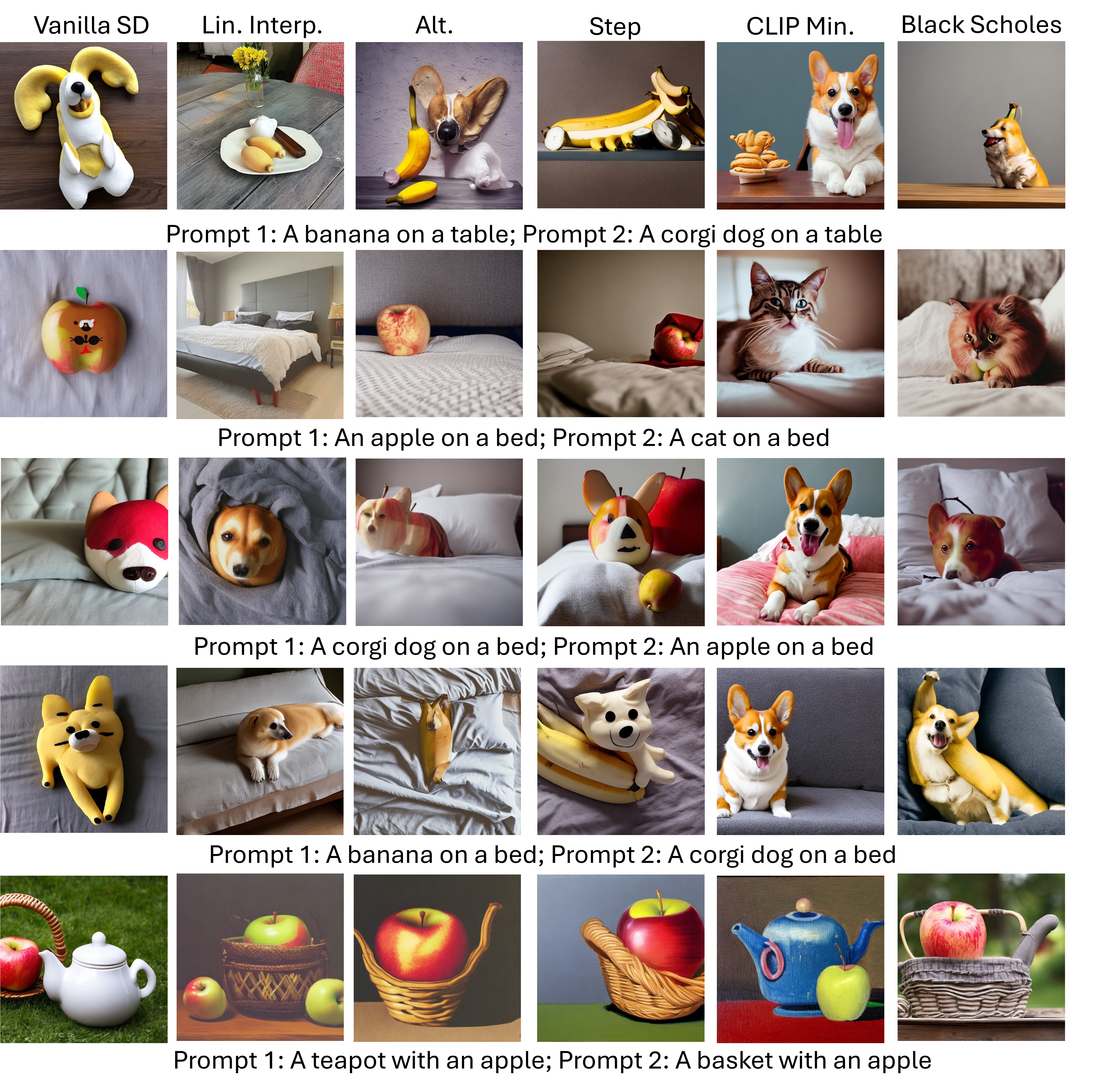}
    \caption{Vanilla stable diffusion (SD) is able to generate images satisfying both text prompts in many cases, however it misses out on the fine-grained characteristics of the individual text prompts. For instance, it misses out on the characteristics of the dog, cat, corgi dog and basket/ teapot mixing. Linear interpolation generates images that are not consistent with the text prompts. CLIP Min. is biased towards one of the text prompts. Alternating sampling and Step wise inference strategies generate images with a lot of artifacts, and the generated images are perceptually implausible in many cases. For instance, in order, the issues are: artifacts in dog/banana and missing characteristics of dog, missing characteristics of cat, the corgi dog in the dog/apple image does not describe the dog as well as our Black Scholes method does, the dog is not well described in the fourth example, the teapot is not well described in the final image. Our method is able to generate images that capture the fine-grained characteristics of objects corresponding to both text prompts, and the generated images look realistic with minimal artifacts. The text prompts are from set 2.}
    \label{fig:supp2}
\end{figure}

\begin{figure}
    \centering
    \includegraphics[width=0.99\textwidth]{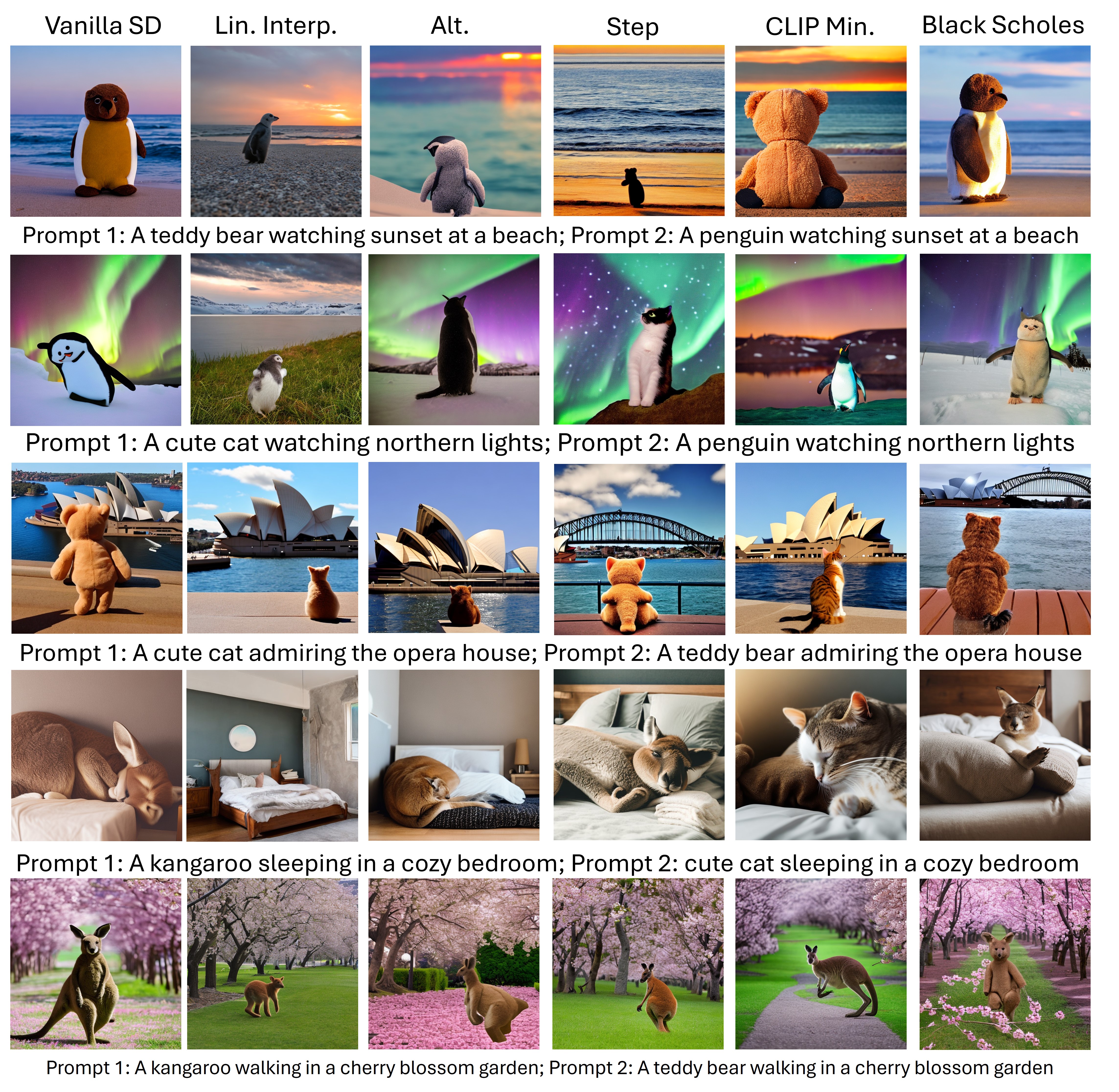}
    \caption{Vanilla stable diffusion (SD) is able to generate images satisfying both text prompts in many cases, however it misses out on the fine-grained characteristics of the individual text prompts. For instance, in the first and fifth image, the characteristics of the teddy bear are missing, the second, third and fourth image do not describe the cat well, Linear interpolation generates images that are not consistent with the text prompts. CLIP Min is again biased towards one of the text prompts. Alternating sampling and Step wise inference strategies generate images with a lot of artifacts, and the generated images are perceptually implausible in many cases. Specifically, the first image does not describe the subjects well, the second and third images are reasonably generated but our Black Scholes method generates a clearer image of the penguin blended cat, and teddy bear blended cat, the results of Alt. for the fourth case has artifacts and Step does not describe the cat well, the teddy bear is not well described in the fifth image. Our Black Scholes method is able to generate images that capture the fine-grained characteristics of objects corresponding to both text prompts, and the generated images look realistic with minimal artifacts. The text prompts are from set 3.}
    \label{fig:supp3}
\end{figure}

\begin{figure}
    \centering
    \includegraphics[width=0.99\textwidth]{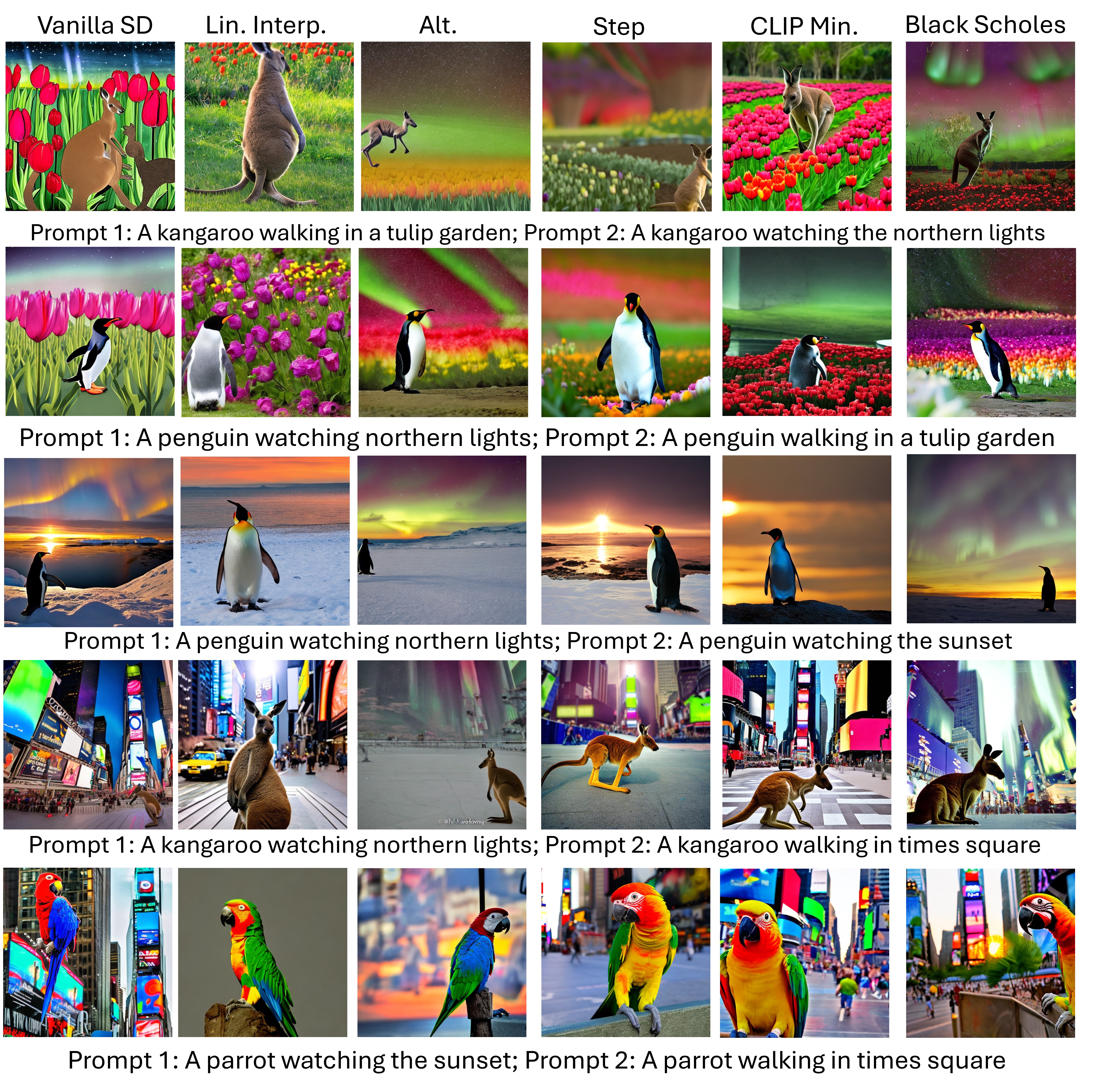}
    \caption{Vanilla stable diffusion (SD) is able to generate images satisfying both text prompts in many cases, however it misses out on the fine-grained characteristics of the individual text prompts. The first two images are cartoon-ish, and Northern lights is not clearly depicted. The third image also misses a clear description of Northern lights. In the fourth image, some of the billboards in Times Square are green, but there is no sign of Northern lights. In the fifth image, the sunset is missing. Linear interpolation generates images that are not consistent with the text prompts. CLIP Min results are biased towards one of the prompts for all images, except the second one. Alternating sampling and Step wise inference strategies generate images with a lot of artifacts, and the generated images are perceptually implausible in many cases. In the first image, the kangaroo is not generated well (almost flying in Alt's result, and the characteristics of the kangaroo are not well generated by Step). In the second image, the characteristics of the tulip garden are not well represented. In the third image, notice how our model generates the characteristics of the sunset and Northern lights better, while ensuring that they are blended well to form a realistic image. In the fourth image, Alt and Step miss out on the characteristics of Times Square and Northern lights respectively. In the fifth image, Alt and Step miss out on the characteristics of Times Square and the sunset respectively. Our Black Scholes method is able to generate images that capture the fine-grained characteristics of objects corresponding to both text prompts, and the generated images look realistic with minimal artifacts. The text prompts are from set 4.}
    \label{fig:supp4}
\end{figure}

\end{document}